\documentclass[letterpaper, 10 pt, conference]{ieeeconf}  % Comment this line out if you need a4paper
\IEEEoverridecommandlockouts                              % This command is only needed if 
\usepackage[font=footnotesize, skip=10pt]{caption}
\usepackage[bb=ams, cal=boondox, scr=cm]{mathalpha}
\usepackage{amsmath}

\usepackage{xspace}
\usepackage{xcolor}
\usepackage{amsfonts}
\usepackage{graphics} % for pdf, bitmapped graphics files
\graphicspath{{figures/}}  % Set the default directory for figures
\usepackage{epsfig} % for postscript graphics files
\usepackage{mathptmx} % assumes new font selection scheme installed
\usepackage{times} % assumes new font selection scheme installed
\usepackage{amsmath} % assumes amsmath package installed
\usepackage{amssymb}  % assumes amsmath package installed
\usepackage{ulem}
\usepackage{siunitx}
\usepackage{verbatim}
\usepackage{booktabs}
\usepackage{multirow}

\allowdisplaybreaks

\usepackage{cite}
\usepackage[colorlinks=true, linkcolor=blue, urlcolor=blue, citecolor=blue]{hyperref}

\makeatletter
\let\origthebibliography\thebibliography
\def\thebibliography#1{%
  \origthebibliography{#1}%
  \scriptsize % \small, \footnotesize, \scriptsize
}
\makeatother

\newcommand{\method}{V-MORALS\xspace}

\newcommand{\edit}[1]{{\color{blue}#1}}
\renewcommand{\edit}[1]{#1}

\title{\LARGE \bf  
V-MORALS: Visual Morse Graph-Aided Estimation \\ of Regions of Attraction in a Learned Latent Space
}

\author{
Faiz Aladin$^1$, Ashwin Balasubramanian$^1$, Lars Lindemann$^{1,2}$, Daniel Seita$^1$% <-this % stops a space
\thanks{$^1$Viterbi School of Engineering, University of Southern California, USA.}
\thanks{$^2$Automatic Control Laboratory, ETH Zürich, Switzerland.}
\thanks{Correspondence to: {\tt\small faladin@usc.edu}}%
}

\begin{document}

\maketitle

\begin{abstract}
Reachability analysis has become increasingly important in robotics to distinguish safe from unsafe states. Unfortunately, existing reachability and safety analysis methods often fall short, as they typically require known system dynamics or large datasets to estimate accurate system models, are computationally expensive, and assume full state information. A recent method, called MORALS, aims to address these shortcomings by using topological tools to estimate Regions of Attraction (ROA) in a low-dimensional latent space. However, MORALS still relies on full state knowledge and has not been studied when only sensor measurements are available. This paper presents Visual Morse Graph-Aided Estimation of Regions of Attraction in a Learned Latent Space (\method). \method takes in a dataset of image-based trajectories of a system under a given controller, and \edit{learns} a latent space for reachability analysis. Using \edit{this} learned latent space, our method is able to generate well-defined Morse Graphs, \edit{from which we can compute ROAs} for various \edit{systems and controllers}. \method provides capabilities similar to the original MORALS architecture without relying on state knowledge, and using only high-level sensor data. 
Our project website is at: \url{https://v-morals.onrender.com}. 
\end{abstract}

\section{Introduction}

With current reachability analysis methods, it is difficult to compute reachable sets for a dynamical system under a given controller when the system is high-dimensional or the controller has a complex structure~\cite{bansal2017hamiltonjacobireachability,mitchell2005time}. To analyze \edit{such dynamical behavior}, we build upon the architecture from~\cite{Vieira2024MORALS} called \textbf{Mo}rse Graph-aided discovery of \textbf{R}egions of \textbf{A}ttraction in a learned \textbf{L}atent \textbf{S}pace (MORALS), to generate a Morse Graph \edit{from which we can compute} the Regions of Attraction (ROA)~\cite{bansal2017hamiltonjacobireachability,tedrake2010lqrtrees} in a learned latent space. Morse Graphs~\cite{Vieira2022MorseGraphs,vieira2022dataefficientcharacterization} provide a powerful way to understand the long-term behavior of complex dynamical systems by building a discrete graph of the system's dynamics, which allows us to analyze safety by approximating its behavior. \edit{The derived ROA} is vital to determining the safety of a system because it indicates if a robot's trajectories will converge to an equilibrium point (i.e., a safe or failure state). 

Applying \edit{a Morse Graph directly} to high-dimensional systems is computationally expensive. %, which is why MORALS uses a low-dimensional latent space to build the Morse Graph and ROA. 
Therefore, MORALS defines a learned latent space by encoding state information into a lower dimension while simultaneously learning the transitions between latent vectors with a latent dynamics network. By generating the Morse Graph and ROA within this lower dimensional space, MORALS provides \edit{an efficient} way to analyze the safety of high-dimensional systems and complex controllers. 

% Daniel: I removed the 'makes temporal prediction straightforward' because if the state is complete yet "high dimensional" that could still make it hard. It could risk someone who works in temporal prediction on states question our claim. 
However, extending this architecture to operate on visual data introduces significant challenges. State representations provide a complete and concise description of a system's configuration, including explicit dynamic variables like joint velocities. In contrast, a single image lacks this motion data, resulting in partial observability and ambiguity, as multiple future states could plausibly follow a single visual frame. Images are also much higher in dimensionality. % than state-based information. 
For example, in CartPole, a standard control task in DeepMind Control Suite~\cite{tassa2018deepmind}, the state has 4 dimensions, while a corresponding image could be orders of magnitude larger in dimension. % much larger in dan image of a CartPole with size 80x170 is roughly 3,400x larger in dimension size. 
When encoding images, latent vectors are unable to store the same level of information as a state-based approach with the same latent dimensions. This also complicates learning dynamics within a latent space~\cite{lippi2024ensembleLSR}, as the transition between two latent vectors is only physically meaningful if the corresponding sequence of reconstructed images~\cite{hafner2019learning} represents a valid progression in the original environment. 

\begin{figure}[t]
    \centering
    \includegraphics[width=1.0\linewidth]{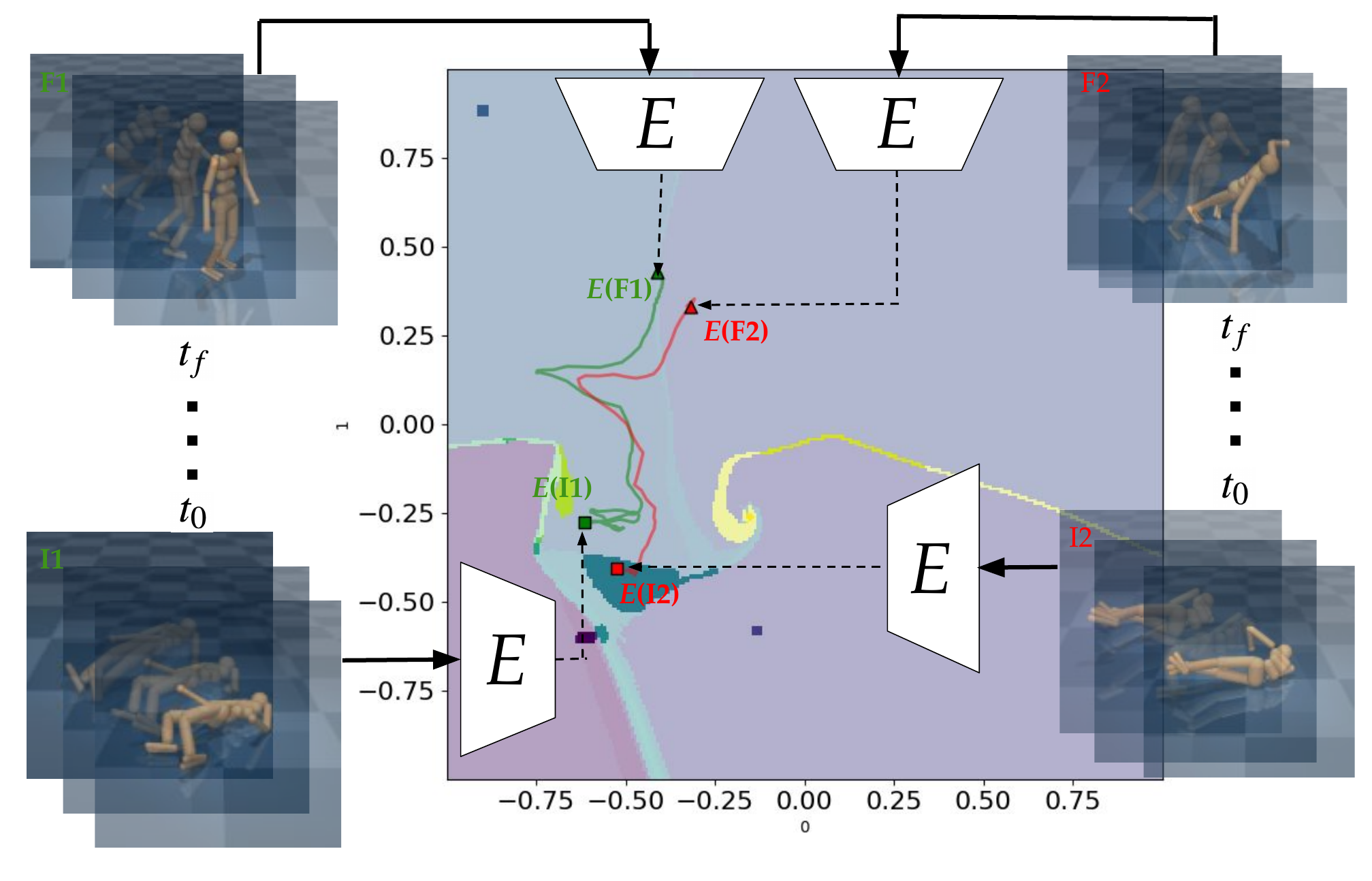}
    \caption{
         Regions of Attraction (ROA) of the GetUp controller~\cite{10.1145/3528233.3530697} on a Humanoid. Given a trajectory of images, we learn a latent space to generate a Morse Graph and ROA. $I1$ and $I2$ are examples of image sequences (with simplified notation to suppress details). Here, $E(\cdot)$ refers to the encoded latent vector of an image sequence. The $t_0$ and $t_f$ represent, respectively, the initial and final images in the trajectory. Each colored region corresponds to a different attractor, allowing the system to predict the long-term outcome of a trajectory, such as success $I1 \rightarrow F1$ or failure $I2 \rightarrow F2$, and providing safety analysis without access to the system's state information.
    }
    \label{fig:pull}
    \vspace{-15pt}
\end{figure}

In this paper, we present \method as a nontrivial extension of MORALS to deal with partial observability. \method learns system dynamics from visual data (see Figure~\ref{fig:pull} for an example). To do so, we preprocess each image to generate a binary mask, which isolates the system from the background to reduce input complexity. To embed temporal information, we encode a sequence of sequential frames into a single latent vector. This conditions the representation on an initial trajectory and thus constrains the possible future states.
We implement this spatiotemporal encoding using a 3D convolutional autoencoder~\cite{tran2015learning}. The encoder compresses the image sequence, capturing its visual content and temporal evolution, while the decoder reconstructs it. Both components are trained jointly with a latent dynamics network. Using these networks, \method effectively captures the underlying dynamics of a controller applied to a system, in a learned latent space. The contributions of this paper include:
\begin{enumerate}
    \item \method, a method which extends MORALS by generating Morse Graphs and ROA in a latent space with only partial observability. Our approach addresses the issues of using high-dimensional images and capturing the dynamics of sequences of images, to define a learned latent space. 
    \item We provide extensive empirical validation of our approach on four standard control benchmarks (Pendulum, CartPole, Humanoid, and Acrobot). Our experiments demonstrate that the model successfully learns the underlying system dynamics and generates accurate Morse Graphs and Regions of Attraction across various controllers and latent space dimensionalities.
\end{enumerate}

\section{Related Work}

\subsection{Reachability and Safety Analysis}

Classical reachability analysis provides formal guarantees about whether trajectories of a dynamical system remain within safe sets or converge to desired equilibria. Tools such as Hamilton-Jacobi reachability have been widely applied in robotics and control to compute forward and backward reachable sets, but can scale poorly due to the exponential dependence on system dimensionality~\cite{bansal2017hamiltonjacobireachability,mitchell2005time,chen2018decomposition,mitchell2007toolbox}. Work such as computing backward reachable tubes using neural networks~\cite{bansal2021deepreachdeeplearning} can help with scaling for some dimensions but training time still increases exponentially.  

Recent work has extended Hamilton–Jacobi reachability to hybrid dynamical systems for walking robots~\cite{choi2022computationregionsattractionhybrid} and to adaptive shielding for safe reinforcement learning in real-world robots~\cite{lu2025safeshieldingHJ}, but these approaches have only been tested on low-dimensional state spaces. 
\edit{Related efforts also explore learning Regions of Attraction for nonlinear systems directly from data~\cite{chen2021learningregionattractionnonlinear} with the help of Lyapunov functions~\cite{richards2018lyapunovneuralnetwork,lederer2019asymptotocstability}, but typically to estimate a single attractor.} 

Complementary approaches based on Control Barrier Functions (CBFs) provide a more tractable alternative, enabling real-time safety filtering of control policies~\cite{ames2019controlbarrierfunctionstheory}. 
Recent advances extend these tools by learning CBFs directly from expert demonstrations~\cite{robey2020CBFfromdemonstrations} and by developing model-free formulations that scale to high-dimensional systems without requiring explicit dynamics models~\cite{oh2025safetyagencyhumancenteredsafety}. 
Nevertheless, these methods still assume access to low-dimensional state representations. In contrast, we explore safety analysis in scenarios where explicit system dynamics and full state information are unavailable, requiring us to rely on images.

\subsection{Latent Space Representations for Planning and Control}

Learning compact latent representations has become a common strategy for enabling robots to reason from high-dimensional sensory inputs such as images. Rather than operating directly in pixel space, these methods train deep generative or predictive models to capture structure in a lower-dimensional latent space, where dynamics are easier to model and manipulate. 
This abstraction has improved planning, control, and reinforcement learning~\cite{hafner2019learning,hafner2021masteringataridiscreteworld,hafner2020dreamcontrollearningbehaviors,hafner2023mastering}. 

In robotics, latent representations have been leveraged to support a variety of planning frameworks. For example, latent spaces have guided motion planning~\cite{itcher2019motionplanninglatent} and enabled graph-based search for long-horizon planning~\cite{lippi2024ensembleLSR,lippi2020lsr,lippi2022aug-conn-explore,lippi2024visualactionplanningmultiple}. 
More recent work has applied latent relational dynamics to multi-object manipulation and rearrangement~\cite{huang2024latentspaceplanning}. 
While these approaches demonstrate the benefit of latent spaces for improving planning and control efficiency, they generally do not \edit{analyze system safety}. In contrast, our work leverages latent representations as the foundation for estimating Regions of Attraction for safety analysis.

\subsection{Safety and Reachability in Latent Spaces}

A growing body of work explores combining safety analysis with latent representations. 
For example,~\cite{lutkus2025latent} propose methods for designing controllers directly in latent spaces with provable stability and safety guarantees, enabling formal reasoning without operating in the full high-dimensional observation space. 
Recent work has also extended reachability analysis to latent spaces to generalize safety beyond collision avoidance, demonstrating that latent models can capture broader classes of constraints relevant for robotics~\cite{nakamura2025generalizing}.

The most closely-related work to ours is MORALS~\cite{Vieira2024MORALS}, which combines latent dynamics with Morse Graph analysis to identify attractors and their corresponding ROAs. MORALS provides a powerful framework for \edit{checking} safety and long-term outcomes, but it assumes access to \edit{state information} and has not been studied in settings where only raw sensory data is available. In contrast, our work extends MORALS to operate under partial observability using image-based inputs, and thus performs safety analysis into domains where only high-dimensional observations are accessible.

% Daniel: trying to make this appear on the top of page 3, so I'm reshuffling the order of the figure LaTeX code. 

\section{Problem Statement and Formulation}
\label{sec:problem_stmt_prelims}

% \section{Preliminaries}
\subsection{Dynamics and Observation Function}
\label{sec:dynamics and observation function}
%We consider a discrete-time dynamical system controlled by an unknown policy and are given a set $\mathcal{S}$ which is the set of all states for a dataset of trajectory rollouts.
\edit{We consider a discrete-time dynamical system with an $n$-dimensional state space $\mathcal{S} \in \mathbb{R}^n$, governed by a known controller.} The system's state $s_t \in \mathcal{S}$ at time $t$ evolves according to the unknown dynamics $f$ such that the next state is $s_{t+1}=f(s_t, u_t)$, where $u_t$ is a control \edit{input} from a given state or image-based controller and with $s_0 \in \mathcal{S}_{initial}$ being the initial state. The underlying controller can be state or image-based, but our method only uses images when performing safety analysis. While the true state $s_t$ is inaccessible, we can observe the system through an observation function, $\phi: \edit{\mathcal{S}} \rightarrow \mathbb{R}^{H\times W\times C}$ in image space, where $H, W$, and $C$ indicate the height, width, and number of channels, respectively. The image observation at time $t$ is thus given by $I_t = \phi(s_t)$.

\subsection{Reachable Sets}
%Our objective is to compute reachable sets using the partial observability set $\mathcal{I}$, defined as $\phi(\mathcal{S})$, and determine whether a set of initial states $\mathcal{I}_{initial}$ will result in desirable or undesirable behavior.
\edit{Our objective is to compute reachable sets having only access to images, and to determine if the set of initial images $\mathcal{I}_{initial}$, such that $s_0 \in \mathcal{S}_{initial}$ and $I_{0} = \phi(s_0)$, will result in a desirable or undesirable behavior.} We compute reachable sets to see how $I_0 \in \mathcal{I}_{initial}$ will converge to an attractor. Conceptually, \edit{we consider the image space dynamics  $I_{t+1}=g(I_t)$ where $g$ encodes  $\phi(f(s_{t}, u_t))$ determined by the given controller.} We also define $r$ as the number of recursive rollout steps of $g$ applied to $I_t$ such that $I_{t+r}=g^r(I_t)$. This also means that $I_t=g^0(I_t)$. Given the set of initial observations $\mathcal{I}_{\text{initial}}$, the reachable set, denoted $\mathcal{R}(\mathcal{I}_{\text{initial}})$, is the set of all observations that can be reached from any initial observation in $\mathcal{I}_{\text{initial}}$ over any number of future time steps. Formally, it is the union of all forward images of the initial set under the dynamics $g: \mathcal{R}(\mathcal{I}_{\text{initial}}) = \bigcup_{r=0}^{\infty} \{ g^r(I_0) \mid I_0 \in \mathcal{I}_{\text{initial}} \}$.
By computing this set, we can partition the observation space into regions corresponding to distinct long-term behaviors, and identify which initial conditions lead to desirable or undesirable outcomes. There is a subtle challenge here due to the use of observations. When we observe states directly (i.e., when the observation map $\phi$ is the identity function), formally defining desirable behaviors is trivial, as it reduces to reaching a goal region or avoiding obstacles. However, defining desirable behavior in image space is not immediate, highlighting a distinct challenge of our work.

\subsection{Problem Formulation}
\label{sec:problem formulation}
We consider a dataset $\mathcal{D} = \{\tau_i\}_{i=1}^N$ containing $N$ trajectories, where each $\tau_i = \{s_0^{(i)}, \ldots, s_k^{(i)} \}$ represents an underlying sequence of states. We do not need to know the system, controller or observation function, or the state information of $\tau_i$ as long as we can collect image-based trajectories.
Each image $I_t^{(i)}$ is generated from its corresponding state $s_t^{(i)}$ via the observation function: 
$\phi(s_t^{(i)}) = I_t^{(i)}$. % Daniel: I don't think this warrants a new line, and it takes a lot of space, so I moved it inline.
We also define: $\mathcal{I}_i = \{I_0^{(i)}, \ldots, I_k^{(i)}\}$,
as the set of images in trajectory $\tau_i$. We denote our image dataset as $\mathcal{D}_I = \{ \mathcal{I}_i \}_{i=1}^N$. \edit{Lastly, we define a set of labels $Y$, such that $Y_i$ denotes the trajectory outcome (success or failure) of $\tau_i$.} We define $Y_i = 1$ for the trajectories that result in successful completion of the task (a desired outcome) and $Y_i = 0$ for those that result in failure (an undesired outcome).

We demonstrate how we can use this image dataset $\mathcal{D}_I$ to analyze the system's \edit{dynamical behavior}. Specifically, we aim to compute the initial set $\mathcal{I}_{initial}$ where image trajectories within the reachable set $\mathbb{R}(\mathcal{I}_{initial})$ have a success label of $Y_i=1$.  Our method \edit{trains a model} to achieve a dual objective: to predict the eventual outcome of a given trajectory (success or failure), and to generate a high-level, combinatorial map of the state transitions within a learned latent space. 

%See Figure~\ref{fig:pipeline} for an overview of the pipeline. 

\section{MORALS Background}
\label{sec:morals_background}

\subsection{Latent Space and Attractors}
\label{sec: latent space and attractors}
To analyze the long-term behavior of high-dimensional systems, the MORALS framework~\cite{Vieira2024MORALS} leverages tools from combinatorial topology. MORALS creates a finite, discrete representation of a system's dynamics, which allows for identifying attractors and their corresponding ROA. To provide context for our contributions \edit{and establish necessary background}, we outline two key components from MORALS that are central to our image-based adaptation: the use of a learned latent space to manage dimensionality, and the application of Morse Graphs to analyze a system's latent dynamics.
We note that MORALS applies to our problem only if the observation function, $\phi$, is the identity. 

% In Section~\ref{sec:dynamics and observation function}, we define $I_t = \phi(s_t)$, which means our observation $I_t$ must be treated as the system's state information in our context. Using our dynamics function $g$, we can effectively apply $g$ to $I_t$ the same way we would apply $f$ to $s_t$. Rather than using a control $u_t$, we can use a sequence of sequential images to represent $u_t$ (outlined in Section~\ref{sec:image based data generation}). 
\begin{figure}[t]
    \centering
    % --- placeholder box: replace with actual graphic later ---
    \includegraphics[width=1.0\linewidth]{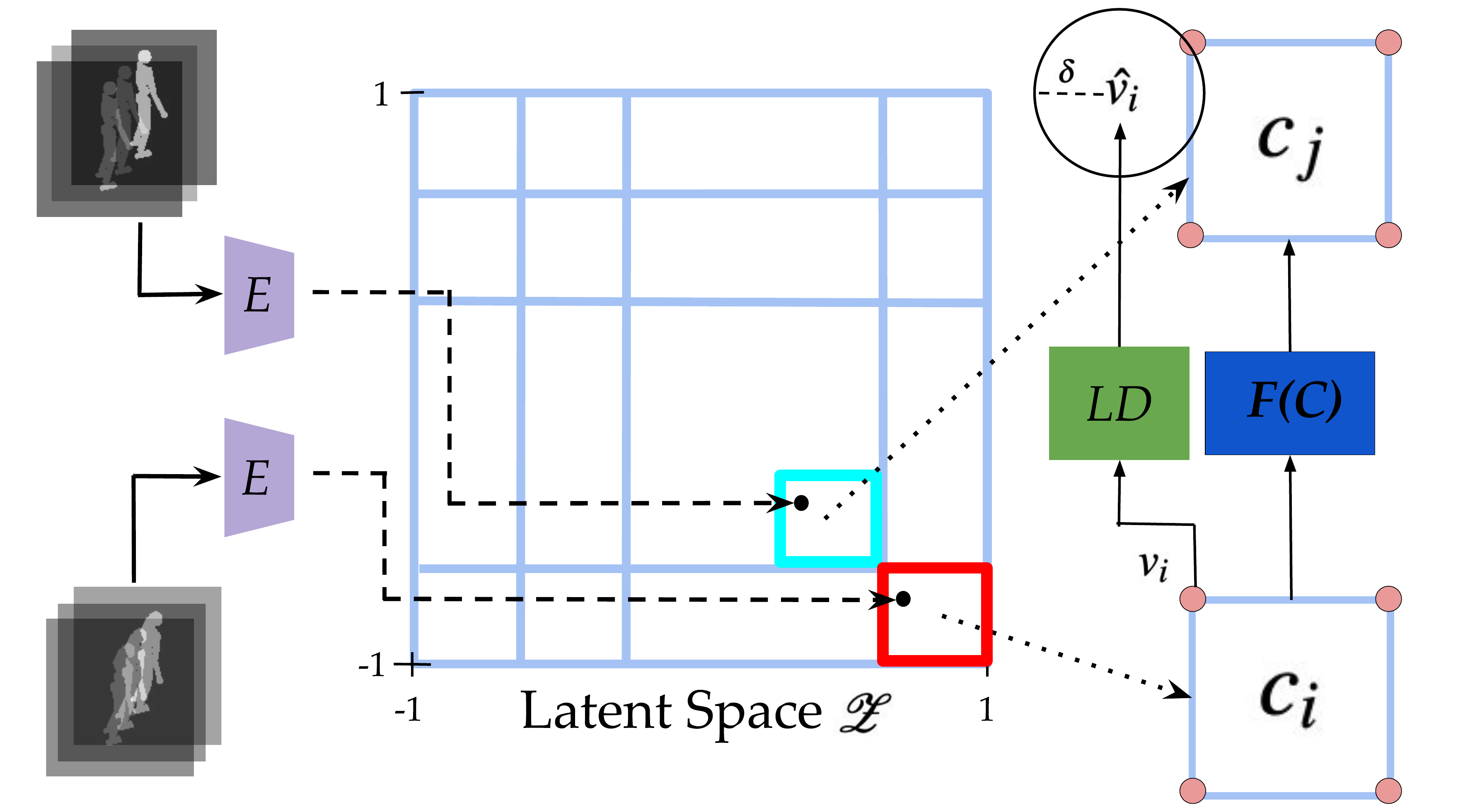}
    \caption{The process for constructing the directed graph \textit{F}, which serves as a combinatorial map of the system's dynamics within the learned latent space $\mathcal{Z}$. In MORALS (Section~\ref{sec:morals_background}), state sequences are first projected into the latent space by the encoder $E$.  \method (Section~\ref{sec:proposed_method}) builds upon this by projecting image sequences into the latent space as shown above. This space is then discretized into a grid of cells $C$. To determine the flow between cells, the corner points of a given cell ($c_i$) are propagated through the learned dynamics network $LD$. A safety bubble with radius $\delta$ is created around each predicted point to account for prediction uncertainty. A directed edge is drawn from $c_i$ to another cell $c_j$ in the graph $F(C)$ if the union of these safety bubbles intersects with $c_j$. This process is repeated for all valid cells to build a complete map of the latent dynamics.}
    \label{fig:ROA_generation}
    \vspace{-10pt}
\end{figure}
%A primary challenge in analyzing complex robotic systems is the high dimensionality of their state spaces. 
\edit{Even when state measurements are available, a key challenge in analyzing dynamical systems is the high state space dimensionality}. MORALS addresses this by learning a low-dimensional latent space, $\mathcal{Z}$, that captures the essential dynamics of the system. This uses an encoder defined as: $E:\mathcal{S} \rightarrow \mathcal{Z}$, that maps the high-dimensional state $s \in \mathcal{S}$ to a low-dimensional latent vector $z \in \mathcal{Z}$, a latent dynamics network $LD: \mathcal{Z} \rightarrow \mathcal{Z}$ that predicts a future latent vector of  $z \in \mathcal{Z}$, and a decoder $D:\mathcal{Z} \rightarrow \mathcal{S}$ that reconstructs the original state from the latent vector. By training the encoder and decoder on system trajectories, the framework obtains a compressed representation where the complex dynamics can be analyzed more efficiently. 

Once the dynamics are learned in the latent space, MORALS employs tools from combinatorial topology to \edit{analyze the behavior of the dynamical system}. MORALS relies on discretizing the learned latent space, which leads to an exponential expansion of the search space as dimensions increase. This limitation forces MORALS to use lower-dimensional state information, but presents a unique challenge for high-dimensional measurements such as images that typically require a higher latent dimension~\cite{hafner2019learning}. The central tool for this is the Morse Graph~\cite{Vieira2022MorseGraphs}, a directed acyclic graph that provides a finite, combinatorial \edit{representation} of the system's dynamics. The Morse Graph's nodes correspond to the system's recurrent sets, and its edges describe the flow between them. We define a recurrent set as a collection of states where the system can cycle indefinitely, meaning that any state in the set is reachable to other states within that set. The leaf nodes of this graph represent the system's attractors, which are stable states or limit cycles to which trajectories converge. By identifying attractors, one can then compute their ROA, which is the set of all initial states that are guaranteed to converge to a specific attractor. We can determine which attractor an initial state will belong to by traversing the Morse Graph starting from the attractor to the initial state. This technique allows MORALS to predict the final outcome (e.g., success or failure) of a trajectory from its initial state, a capability that our work aims to extend to systems observed only through images.

\subsection{Morse Graph and ROA Generation}
\label{sec: morse graph and regions of attraciton}

Using the $d$-dimensional learned latent space $\mathcal{Z}$, we can construct a Morse Graph $MG$, which provides a discrete representation of the system's global behavior, and its corresponding ROA. We first discretize this latent space, which is bounded by $[-1,1]^d$, into a grid of hypercubes, or ``cells" (see Figure~\ref{fig:ROA_generation}). We then sample sequences of states from the trajectories in the training data, and encode each of them into $\mathcal{Z}$ using the trained encoder, $E$. To ground our analysis in physically meaningful states, we focus only on the cells that contain these encoded data points and their immediate neighbors. We denote this collection of valid cells as $C$. % See Figure~\ref{fig:ROA_generation} for a visualization.

To build $MG$, we determine the transitions between cells in $C$ using the trained latent dynamics model $LD$. This involves creating a directed graph, $F$, where each cell is a node. An edge connects cell $c_i$ to another cell $c_j$ if $LD$ can cause a transition from a state in $c_i$ to a state in $c_j$. However, since it is computationally infeasible to calculate the future state for every point in a given cell $c_i \in C$, we only consider the corner points in $c_i$ and denote these points as $V$. For a corner point $v_i \in V$, we apply our latent dynamics network to give $LD(v_i) \rightarrow \hat{v_i}$ for $r$ rollout steps. The number of rollout steps is chosen based on the length of trajectories for the system (i.e longer trajectories require more rollout steps while shorter ones require fewer steps).  To account for the uncertainty in this prediction and to ensure that we capture all possible future states for any point within a cell $c_i$, a $\delta$-closed ball is created around $\hat{v_i}$ and each predicted corner point. The radius of this ball is determined by the Lipschitz constant $L$ of the dynamics, which sets an upper bound on the maximum possible divergence on trajectories. A transition from cell $c_i$ to cell $c_j$ is considered possible if the union of these $\delta$-closed balls of each predicted corner point intersects with $c_j$. This process is repeated for all valid cells, resulting in the directed graph $F$, which serves as a combinatorial map of the system's dynamics.
%reliable, combinatorial map of the system's dynamics.

With $F$ constructed, we now simplify this detailed map of cell-to-cell transitions into the more abstract and interpretable Morse Graph, $MG(F)$. This step distills the complex, cyclical dynamics within $F$ into their essential, long-term behaviors.
The construction of $MG(F)$ begins by decomposing the graph $F$ into its Strongly Connected Components (SCCs). An SCC is a subgraph where every node, representing a cell in the latent space, is reachable from every other node within that same subgraph. These SCCs represent the recurrent sets of the dynamics, which are regions where the system can cycle indefinitely. 
Each identified SCC is then collapsed into a single node (i.e a Morse Set) in a new graph. A directed edge is drawn from one Morse Set to another if there exists at least one edge in the original graph $F$ from a cell in the first SCC to a cell in the second. The resulting graph is the Morse Graph, $MG(F)$. By containing cycles within nodes, $MG(F)$ is constructed as a directed acyclic graph representing the state transition hierarchy of $\mathcal{Z}$.

\begin{figure}[t]
    \centering
    % --- placeholder box: replace with actual graphic later ---
    \includegraphics[width=.9\linewidth]{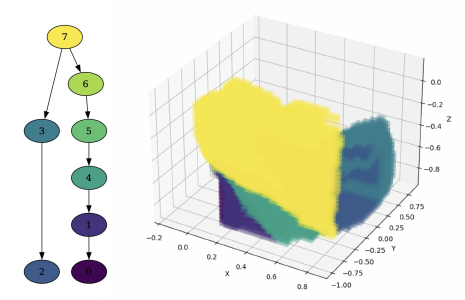}
    \caption{Morse Graph and ROA of the Get Up controller applied to a Humanoid, with a latent space dimension of 3. The colors for each Morse Node describe how the set of states changes between each node. The dark blue node represents the attractor of the success region while the dark purple node represents the attractor of the failure. \method can successfully analyze a complex, high-degree-of-freedom system using only high-dimensional image data, providing an interpretable, low-dimensional map to predict bistable tasks. Additionally, \method further extends MORALS by being able to visualize ROAs in a 3 dimensional space. 
    }
    \label{fig:result3}
    \vspace{-10pt}
\end{figure}

We then use our Morse Graph to \edit{derive the ROA} for any given attractor, \edit{which corresponds to a leaf node in the graph} (see Figure~\ref{fig:result3}). The ROA of an attractor is then defined as the set of cells in $C$ that have a path in $F$ leading to any of the cells within the SCC corresponding to that attractor. By computing these ROAs, we can partition the state space into distinct basins, providing a formal guarantee on the long-term outcome for any initial state within those regions.

\begin{figure*}[t]
    \centering
    % --- placeholder box: replace with actual graphic later ---
    \includegraphics[width=1.0\linewidth]{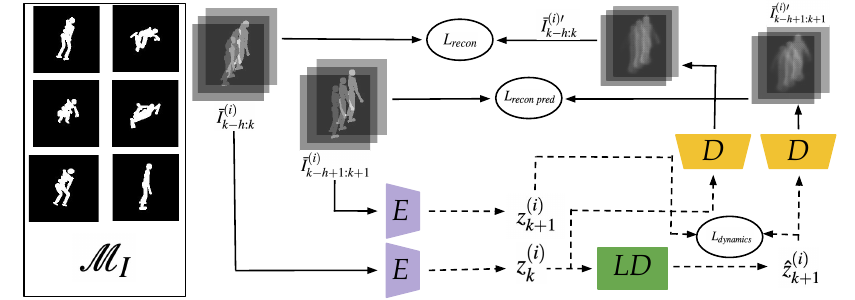}
    \caption{The \method architecture and training pipeline. To the left, we show representative binary images from our dataset $\mathcal{M}_I$ used to form training samples (from the Humanoid task). Using $\mathcal{M}_I$, we randomly sample a sequence of input images to form a training sample, $\bar{I}_{k-h:k}$. This is mapped to a low-dimensional latent vector $z_k$ by an Encoder ($E$). The dotted arrow lines represent a visualization of the operations in latent space $\mathcal{Z}$. A latent dynamics network ($LD$) is trained to predict the future latent state $\hat{z}_{k+1}$. A Decoder ($D$) reconstructs the image sequence $\bar{I}_{k-h:k}^{\prime}$ from the latent state $z_k$.}
    %See Section~\ref{sec:proposed_method} for more details.}
    \label{fig:pipeline}
    \vspace{-10pt}
\end{figure*}

\section{Proposed Method: \method}
\label{sec:proposed_method}

\subsection{Image-Based Data Generation}
\label{sec:image based data generation}

For each image in our dataset $\mathcal{D}_I$, we apply a binary mask to isolate the system from the background. This is a crucial preprocessing step because the dynamics of our chosen tasks are governed entirely by the system's physical configuration (i.e., the positions and angles of its parts). The mask removes dynamically irrelevant information, such as texture and lighting, enabling the model to learn a more robust and accurate representation of the system's state. Formally we define this preprocessing function as $\psi: \mathbb{R}^{H\times W\times C} \rightarrow \mathbb{B}^{H\times W}$ in the binary output space where $H$ is the height and $W$ is the width. In this space, the number of channels $C$ is set to 1. We can then define the binary mask dataset as ${\mathcal{M}_I= \{\psi(I) \mid I \in \mathcal{D}_I\} }$. 
% \daniel{A reader might question this writing: wouldn't this binary mask risk obscuring many details (depending on the task)?}.

A key challenge, not present in MORALS~\cite{Vieira2024MORALS}, is capturing the unique states of the system using only image frames. A single image frame can correspond to multiple states, creating ambiguity. To capture temporal dynamics, we encode short image sequences, into latent vectors. From each binary image trajectory, % $\bar{\mathcal{I}}_i \in \mathcal{M}_I$, 
we construct a dataset of ordered pairs for training. Each pair consists of two image sequences separated by a time interval. The time interval is a hyperparameter that can be adjusted depending on the task. Let the sequence size (history length) be denoted by $h$. A single training pair is formed by a sequence starting at time $k$ and the subsequent sequence starting at time $k + 1$:$(\bar{I}_{k-h:k}^{(i)}, \bar{I}_{k-h+1:k+1}^{(i)})$,
where a binary image sequence $\bar{I}_{k-h:k}^{(i)}$ includes the following $h$ consecutive images: $\bar{I}_{k-h:k}^{(i)} = \{\bar{I}_{k-h}^{(i)},\bar{I}_{k-h+1}^{(i)}\ldots,\bar{I}_k^{(i)}\}$.
Each trajectory % $\bar{\mathcal{I}}_i$ 
is assigned a binary outcome label, ${Y_i \in \{0, 1\}}$, corresponding to its final behavior. The ordered pair with its label is denoted as: $(\bar{I}_{k-h:k}^{(i)}, \bar{I}_{k-h+1:k+1}^{(i)}, Y_i)$.

\subsection{Model Architecture}

We define a $d$-dimensional latent space $\mathcal{Z} \subseteq [-1,1]^d$ using three different networks: the encoder, decoder, and the latent dynamics network. Our encoder and decoder enable meaningful latent representations while our latent dynamics network reasons about transitions between states in $\mathcal{Z}$.

Our \textbf{encoder}, $E$, is a 3D convolutional autoencoder~\cite{Ravi2020PyTorch3D}, designed to process sequences of images. It maps a binary image sequence $\bar{I}_{k-h:k}^{(i)}$ to a low-dimensional latent vector $z_k \in \mathcal{Z}$, or more formally:  $z_k^{(i)}=E(\bar{I}_{k-h:k}^{(i)})$. Using 3D convolutions is crucial because they operate across spatial (height, width, channel) and temporal (the sequence dimension) axes. Since we use binary masks, each image in the sequence only has 1 channel.  This allows the network to directly learn spatiotemporal features, such as motion and velocity, from the raw pixel data.
To enforce the boundaries of the latent space and employ normalization, the final layer of the encoder utilizes a $\tanh$ activation function, ensuring all components of the latent vector $z_k^{(i)}$ are mapped to the range $[-1, 1]$. The encoder is trained to distill the position of the system and the essential dynamical information from the image sequence into this compact vector.

Our \textbf{decoder}, $D$, maps a latent vector $z_k \in \mathcal{Z}$ from the low-dimensional latent space back to the high-dimensional observation space, reconstructing the original image sequence. Architecturally, the decoder mirrors the encoder and uses 3D transposed convolutions~\cite{dumoulin2018guideconvolution} to upscale a latent vector back into a full-resolution image sequence. This function is defined as: $D(z_k^{(i)}) = \bar{I}_{k-h:k}^{(i)\prime}$, where $\bar{I}_{k-h:k}^{(i)\prime}$ is the reconstructed image sequence. Given observations in $\mathbb{B}^{H \times W}$, we first normalize the raw pixel intensities. To guarantee that the reconstructed image sequence $\bar{I}_{k-h:k}^{(i)\prime} \in \mathbb{B}^{H \times W}$, we apply a sigmoid activation function to the final layer of the decoder. The decoder attempts to reconstruct the original input ($ \bar{I}_{k-h:k}^{(i)\prime} \approx \bar{I}_{k-h:k}^{(i)}$), which encourages the latent vector $z_k$ to retain salient information.

Lastly, we define our \textbf{latent dynamics network}, $LD$, as a feedforward neural network. This network operates in $\mathcal{Z}$ and is trained to predict the next latent state based on the current one: $LD(z_k^{(i)}) = {\hat{z}_{k+1}^{(i)}}$, where $z_k^{(i)}$ is the current latent state and $\hat{z}_{k+1}^{(i)}$ is the predicted latent state at the next time step. To ensure the predicted vector remains within the bounds of the latent space, this network also employs a $\tanh$ activation function on its final layer. This allows us to simulate and predict the system's behavior within the latent space. % For more details, we refer the reader to MORALS~\cite{Vieira2024MORALS}.  % Daniel: I think this was added before our recent change to move more MORALS stuff to a background section. 

\subsection{Training Objectives}

To jointly train the model, our total loss function is composed of four components, which are calculated for each training pair: $(\bar{I}_{k-h:k}^{(i)}, \bar{I}_{k-h+1: k + 1}^{(i)}, Y_i)$.

The first component is the autoencoder reconstruction loss, $L_{recon}$. This loss ensures that the encoder-decoder can compress and reconstruct a binary image sequence:  % . The loss is defined as:
\begin{equation}
\label{eq:recon}
L_{recon} = \text{BCE}(\bar{I}_{k-h:k}^{(i)}, D(E(\bar{I}_{k-h:k}^{(i)}))),
\end{equation}
where we use the Binary Cross-Entropy (BCE) loss to measure error on our reconstructed binary image sequences. 

Second, we compute the latent dynamics loss $L_{dynamics}$. This loss ensures that the difference between our output of the latent dynamics network and our encoded sequence ${z_{k+1}^{(i)}}$ is minimized. This loss is formally defined as:
\begin{equation}
\label{eq:dynamics}
L_{dynamics} = \text{MSE}(E(\bar{I}_{k-h+1: k + 1}^{(i)}), LD(E(\bar{I}_{k-h:k}^{(i)}))).
\end{equation}
We use the Mean-Squared Error (MSE) loss to minimize the Euclidean distance between the two latent vectors. % in $\mathcal{Z}$.

Third, we compute the reconstruction loss $L_{recon\;pred}$ where we reduce the loss between the second sequence in the pair, $\bar{I}_{k-h+1: k + 1}^{(i)}$ and the reconstruction of our latent prediction:
\begin{equation}
\label{eq:recon_pred}
L_{recon\;pred} = BCE(\bar{I}_{k-h+1: k + 1}^{(i)}, D(LD(E(\bar{I}_{k-h:k}^{(i)})))).
\end{equation}
Similar to Equation~\ref{eq:recon}, we use BCE to minimize the loss between the two sequences of binary images.

Lastly, we define a contrastive loss, $L_{contrast}$, that structures the latent space by training the model to group latent vectors with the same $Y_i$. This loss operates on a group of latent vectors from the positive $\mathcal{Z}_{pos}= \{ z \in \mathcal{Z} \mid Y_i = 1 \}$ and negative ($\mathcal{Z}_{neg}=\{ z \in \mathcal{Z} \mid Y_i = 0 \}$) classes, with two objectives:
%total_loss = inter_class_loss.mean() + intra_class_loss_pos.mean() + intra_class_loss_neg.mean()
%Original MORALS uses interclass loss and the change we make is to add intra class loss as well too 
\begin{enumerate}
    \item Inter-Class Loss: This component pushes the two clusters apart. It penalizes any pair of positive and negative latent vectors that are closer to each other than a defined margin, $m$.
    \item Intra-Class Loss: This component makes each cluster tighter. It penalizes the pairwise distances between all vectors within the positive cluster and, separately, within the negative cluster.
\end{enumerate}
This loss can be defined as: 
\begin{equation}
\label{eq:contrast}
\begin{split}
L_{contrast} = &\sum_{z_p \in \mathcal{Z}_{pos}, z_n \in \mathcal{Z}_{neg}} \max(0, m - \|z_p - z_n\|_2^2) + \\
&\quad \sum_{z_{p_i}, z_{p_j} \in \mathcal{Z}_{pos}} \|z_{p_i} - z_{p_j}\|_2^2 + \sum_{z_{n_i}, z_{n_j} \in \mathcal{Z}_{neg}} \|z_{n_i} - z_{n_j}\|_2^2.
\end{split}
\end{equation}
The first term represents the inter-class loss, while the remaining two terms correspond to the intra-class loss for their respective classes. We extend the MORALS framework by adding intra-class loss to organize the latent space and help the model differentiate better between successful and failure trajectories via clustering. Using these four components, we define our total loss function as follows:
\begin{equation}
\label{eq:total_loss}
L_{total} = \lambda_1 L_{recon} + \lambda_2 L_{dynamics} + \lambda_3 L_{recon\;pred} + \lambda_4 L_{contrast},
\end{equation}
where $\lambda_{1}$ through $\lambda_{4}$ are weights chosen to ensure high-fidelity reconstructions from the latent space, while also learning an accurate model of the system's dynamics in $\mathcal{Z}$. Using this learned latent space, we can compute the Morse Graph and ROA for the system (see Section~\ref{sec: morse graph and regions of attraciton}).

\begin{comment}
Our method is not variational~\cite{kingma2014VAE} as we found that it was not necessary to achieve promising results. \edit{The stochastic nature of a variational autoencoder introduces nontrivial challenges for our analysis. Each input would map to a distribution of latent states, increasing the number of potential transitions between cells. This results in a Morse Graph with significantly more nodes and complexity, undermining its utility for a high-level, interpretable analysis.} 
\end{comment}
%However in future extensions of \method, we will consider such approaches \edit{and try to mitigate this issue}. 

% Daniel note: this was an earlier comment but let's still keep this in mind. 
% \daniel{We need to discuss the rationale of each loss function. Why does it help? We also need to clarify anything that might be new compared to MORALS. Finally, we also need to clarify that we add all these according to some hyperparameter (and how we select that).}

% Daniel: the VAE paper here is the standard one people use for variational latent spaces. We can potentially say this in the conclusion or future work. 

\subsection{Simulation Tasks}

We evaluate our method through four tasks: Pendulum, CartPole, Acrobot and Humanoid, each with a distinct stabilization goal. The Pendulum's objective is to swing up and balance, the CartPole must balance its pole while moving to the center, the Acrobot must swing up over its base, and the Humanoid must maintain a standing posture. We select these tasks due to the inherent bistable characteristics of the systems. 
Predicted trajectory rollouts can therefore be easily classified as either success or failure.

\section{Experiments}
\label{experiments}
% Daniel: moving the figure earlier since we reference it earlier (slightly before the experiments section). 

To get image data, we use the existing MORALS dataset~\cite{Vieira2024MORALS} and render trajectories in MuJoCo~\cite{todorov2012mujoco}. The Humanoid images are rendered using a public repository~\cite{10.1145/3528233.3530697}. For Pendulum and CartPole, we use an LQR controller~\cite{anderson1990optimal}. For the Acrobot task, we used a DDPG vision based controller~\cite{yarats2021drqv2}. We also used this controller to collect data for another CartPole to compare the difference between state and vision based controllers. For Humanoid, we use a trained SAC~\cite{haarnoja2018softactorcritic} policy. We use image and state based controllers to show our method can generalize to different kinds of controllers. Regardless of the controller, \method only relies on images. \edit{Each trajectory in Humanoid, CartPole, vision-based CartPole, Acrobot, and Pendulum has 200, 1000, 500, 300 and 20 frames respectively.}
 
\subsection{Experiment Protocol}
\label{experiment protocol}
We evaluate the effect of latent space dimensionality in three stages. 
First, we train two models per system (i.e., task), with latent dimensions of 2 and 3. 
We trained the models for each system for 5000 epochs using the Adam optimizer~\cite{kingma2015adam} with a learning rate of $1 \times 10^{-4}$ on one NVIDIA V100 GPU. Each system's data was partitioned into 80\% for training and 20\% for validation.
% From a dataset of trajectories, where each trajectory for the Humanoid, CartPole, and Pendulum environments consists of 200, 1000, and 20 frames respectively, 
We use the validation dataset to define a single trajectory instance as a tuple. To simplify notation we define $l=len(\bar{I}^{(i)})$ such that the tuple is defined as: $({\bar{I}^{(i)}}_{0:h}, {\bar{I}^{(i)}}_{l-h:l}, Y_i)$, consisting of the initial image sequence, the final image sequence, and the outcome label ($Y_i=1$ for success, $Y_i=0$ for failure). Using these tuples, we create four sets. Let $B_s = \{E({\bar{I}^{(i)}}_{0:h})\ \mid Y_i=1\}$ be the set of initial latent vectors, obtained by encoding the first $h$ frames of every successful trajectory. We also define $B_f = \{E({\bar{I}^{(i)}}_{0:h})\ \mid Y_i=0\}$ to be the set of initial latents in each unsuccessful trajectory. Let $L_s = \{E({\bar{I}^{(i)}}_{l-h:l}) \mid Y_i=1\}$ be the set of successful final latent vectors, obtained by encoding the last $h$ frames of successful trajectories only. Similarly we define $L_f= \{E({\bar{I}^{(i)}}_{l-h:l}) \mid Y_i=0\}$. 

We use the training data to generate a Morse Graph and corresponding ROA of each system, under both latent dimensions, by simulating the learned dynamics for 12 rollout steps. We found that 12 was the minimum rollout steps required to generate a readable Morse Graph (i.e., with few attractors) within reasonable computation time.  We then label the attractor basins in the graph: the attractor containing the latent vectors from $L_s$ is designated the ``success region,'' while all other attractors are designated as ``failure regions'' (see Figure~\ref{fig:result5} in the Appendix for a visualization of the Morse Graph and ROA for the CartPole environment). In cases where $L_s$ and $L_f$ shared the same attractor, the label of the attractor was determined by the higher quantity of latent vectors between the two sets.

%(see the Appendix of our arXiv for a visualization of the Morse Graph and ROA for the CartPole environment).

Finally, we calculate the precision, recall, and F-score by classifying the initial states using our validation data. A prediction for an initial state is correct if \edit{its latent vector in $B_s$ converges to the success region or if its latent vector in $B_f$ converges to a failure region.} The total accuracy of our model is the fraction of its correctly classified initial states. 

%\begin{enumerate}
    %\item Its latent vector in $B_s$ maps to the success region.
    %\item Its latent vector in $B_f$ maps to a failure region.
%\end{enumerate}

\subsection{Results}

% Daniel: no need for this opening paragraph now.
% See Table~\ref{tab:morals_evaluation} for our results. \daniel{Describe the results}

\begin{table}[t]
\centering
\begin{tabular}{@{}lcccc@{}}
\toprule
\textbf{Task} & \textbf{Latent Dim} & \textbf{Precision} & \textbf{Recall} & \textbf{F-score} \\ \midrule
\multirow{2}{*}{Humanoid (SAC) } & 2 & 0.9091 & 0.3846 & 0.5405 \\
& 3 & 1.0000 & 0.7253 & 0.8408 \\ \midrule
\multirow{2}{*}{CartPole (LQR)} & 2 & 0.2917 & 0.2979 & 0.2947 \\
& 3 & 1.0000 & 0.6809 & 0.8101 \\ \midrule
\multirow{2}{*}{CartPole (DDPG)} & 2 & 0.7273 & 0.1441 & 0.2406 \\
& 3 & 0.9870 & 0.7308 & 0.8398 \\ \midrule
\multirow{2}{*}{Acrobot (DDPG)} & 2 & 1.000 & 0.4855 & 0.6537 \\
& 3 & 0.9700 & 0.7029 & 0.8151 \\ \midrule
\multirow{2}{*}{Pendulum (LQR)} & 2 & 0.4118 & 0.8750 & 0.5600 \\
& 3 & 1.0000 & 0.4667 & 0.6364 \\ \bottomrule

\end{tabular}
\caption{Performance of \method in classifying trajectory outcomes across different environments and latent space dimensionalities. For all three metrics (Precision, Recall, and F-score), higher is better.}
\label{tab:morals_evaluation}
\vspace{-5pt}
\end{table}

\begin{table}[t]
\centering
\begin{tabular}{@{}llccc@{}}
\toprule
\textbf{Task} & \textbf{Model} & \textbf{Precision} & \textbf{Recall} & \textbf{F-score} \\
\midrule
\multirow{2}{*}{Pendulum (LQR)} & MORALS & 0.9400 & 0.8500 & 0.8900 \\
 & \method (Ours) & 0.4118 & 0.8750 & 0.5600 \\
\midrule
\multirow{2}{*}{CartPole (LQR)} & MORALS & 0.8600 & 0.7700 & 0.8100 \\
 & \method (Ours) & 0.2917 & 0.2979 & 0.2947 \\
\midrule
\multirow{2}{*}{Humanoid (SAC)} & MORALS & 0.9100 & 0.9100 & 0.9100 \\
 & \method (Ours) & 0.9091 & 0.3846 & 0.5405 \\
\bottomrule
\end{tabular}
\caption{Comparison of \method against the original state-based MORALS framework~\cite{Vieira2024MORALS} where both use a latent dimension of two. For Precision, Recall, and F-score metrics, higher is better.}
\label{tab:comparison_morals}
\vspace{-5pt}
\end{table}

The performance of our \method framework in classifying trajectory outcomes is detailed in Table~\ref{tab:morals_evaluation}. The results demonstrate a \edit{direct} relationship between the dimensionality of $\mathcal{Z}$ (i.e., 2 to 3) and the model's predictive accuracy across all test environments.
% A primary finding is the universal performance improvement when increasing the latent dimension from 2 to 3. 
In CartPole, the F-score increased substantially from 0.2947 to 0.8101. We also observe a similar gain in Humanoid, from 0.5405 to 0.8408. This trend suggests that a 2-dimensional latent space is insufficient to capture the complexity of the system dynamics required for accurate outcome prediction, whereas a 3-dimensional space provides a richer and more effective representation. 

\begin{figure}[t]
    \centering
    % --- placeholder box: replace with actual graphic later ---
    \includegraphics[width=1.0\linewidth]{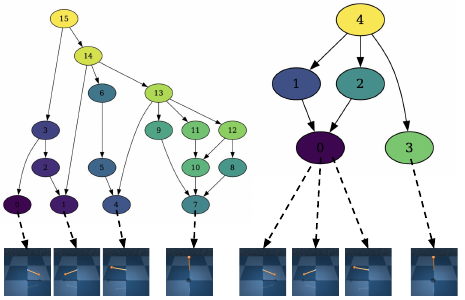}
    \caption{A comparison of Morse Graphs generated for Pendulum using different latent space dimensions. (Left) A 2-dimensional latent space yields a complex graph with multiple attractors (leaf nodes), failing to capture the true bistable nature of the task. (Right) Increasing the latent dimension to 3 results in a simpler and more accurate topological representation. The dynamics correctly converge to two distinct attractors (dark purple and green), which correspond to the task's success and failure modes.}
    \label{fig:result4}
    \vspace{-10pt}
\end{figure}

Qualitatively, we find that the Morse Graph is simpler (with fewer nodes), and encompasses the bistable nature of the task (see Figure~\ref{fig:result4}), when increasing the latent dimension. The magnitude of this improvement also correlates with trajectory length. \method significantly improved in predicting outcomes of initial states for CartPole, which has 1000 frames per trajectories, after increasing the dimensionality. In contrast, \method saw a much smaller gain with the Pendulum environment, which only has 20 frames per trajectory. This indicates that longer trajectories better leverage the increase in latent dimensions. We also find that \method had similar results in classifying trajectories from state-based versus vision-based controllers for CartPole, suggesting that our method generalizes across controllers. 
% Daniel note on wording: use 'significantly' or 'substantially' and not 'dramatically'.

\begin{table}[t]
\centering
\begin{tabular}{@{}lcccc@{}}
\toprule
\textbf{Model} & \textbf{Latent Dim} & \textbf{Precision} & \textbf{Recall} & \textbf{F-score} \\
\midrule
\multirow{2}{*}{MORALS (State-based)} & 2 & 0.9100 & 0.9100 & 0.9100 \\
 & 3 & 0.8900 & 1.0000 & 0.9400 \\
\midrule
\multirow{2}{*}{\method (Ours)} & 2 & 0.9091 & 0.3846 & 0.5405 \\
 & 3 & 1.0000 & 0.7253 & 0.8408 \\
\bottomrule
\end{tabular}
\caption{Performance comparison on the Humanoid (GetUp) benchmark~\cite{10.1145/3528233.3530697} between the state-based MORALS and our image-based \method across latent dimensions.}
\label{tab:humanoid_comparison}
\vspace{-10pt}
\end{table}

% \begin{figure}[t]
%     \centering
%     % --- placeholder box: replace with actual graphic later ---
%     \includegraphics[width=1.0\linewidth]{figures/Figure_6.pdf}
%     \caption{Analysis of the CartPole system's dynamics in a 3-dimensional learned latent space. (Left) The generated Morse Graph identifies three leaf nodes. The images below show the corresponding physical outcomes: the desired balanced state (green node) and two distinct failure modes (light purple, pole falling left; dark purple, pole straight down). (Right) The ROA partition the latent space, where each colored region represents the set of initial states guaranteed to converge to its respective attractor.}
%     \label{fig:result5}
%     \vspace{-10pt}
% \end{figure}

To contextualize the performance of our image-based method, Table~\ref{tab:comparison_morals} provides a direct comparison between \method (using a 2D latent space) and the original MORALS framework which operates on true state information. The performance at a latent dimension of 2 establishes a clear baseline, demonstrating that while the task is inherently more difficult, our approach is viable. This motivates our results in Table~\ref{tab:humanoid_comparison}, which show that this performance gap can be reduced by increasing the latent space dimensionality to better capture the system's complex dynamics. For better visualizations of ROA for each system, we refer the reader to our website: \url{https://v-morals.onrender.com}.
% Daniel note: don't say 'explored in the previous section' say 'as explored in Section~\ref{...}'.

We also evaluate \method when Gaussian noise is introduced to the image observations. We trained two models on the Humanoid task with latent space dimensions of 2 and 3 using the noisy image sequences. Following the evaluation procedure outlined in Section~\ref{experiment protocol}, we assessed the performance of our approach under these noisy conditions. Compared to our models without noise, we observed a performance drop in the F-score for both models, decreasing to 0.2571 and 0.2982 for latent dimensions 2 and 3 respectively. This performance drop can be attributed to our decoder being unable to reconstruct images effectively with noise. 

% To mitigate this issue, we can add an edge filter in our pre-processing step to preserve the outline of the humanoid while removing noise.

\section{Limitations}

% Daniel: do these make sense?
While a promising approach, \method has some limitations that motivate future work. First, our method relies on images being a relatively complete representation of the system, and may struggle with significant partial observability. Additionally, \method requires images to be binarized, potentially omitting essential details within the environment. Second, our method, as with the original MORALS, assumes that there are fixed Regions of Attraction, which might not exhaustively characterize all robotics tasks.  Finally, we only test on a set of simulated tasks, and in future work we plan to test this using images from real-world robotics tasks, as well as to explore different types of manipulators to understand the cross-embodiment transfer~\cite{bauer2025latentactiondiffusion} of latent space analysis. 

\section{Conclusion}

This paper presents Visual Morse Graph-Aided discovery of Regions of Attraction in a learned Latent Space (\method). \method provides capabilities similar to the original MORALS architecture without relying on state knowledge, or the controller used for the system. Our method is able to generate well-defined Morse Graphs and ROAs from different controllers for the Pendulum, CartPole, and Humanoid environments. We hope that this work inspires future research in the analysis of reachability and safety of controllers for complex, high-dimensional scenarios.

\section*{Acknowledgments}
We thank Aravind Sivaramakrishnan and Sumanth Tangirala for their insights on Morse Graph and ROA generation, and for detailing their data collection process.

\bibliographystyle{IEEEtran}
\bibliography{main}

\clearpage
\appendices

\section{Success and Failure Regions of Cartpole}
\label{app:latent_dims}

\begin{figure}[h]
    \centering
    % --- placeholder box: replace with actual graphic later ---
    \includegraphics[width=1.0\linewidth]{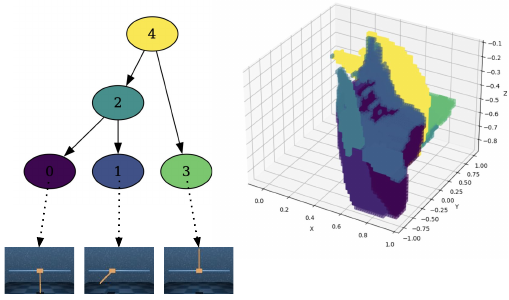}
    \caption{Analysis of the CartPole system's dynamics in a 3-dimensional learned latent space. (Left) The generated Morse Graph identifies three leaf nodes. The images below show the corresponding physical outcomes: the desired balanced state (green node) and two distinct failure modes (light purple, pole falling left; dark purple, pole straight down). (Right) The ROA partition the latent space, where each colored region represents the set of initial states guaranteed to converge to its respective attractor.}
    \label{fig:result5}
    \vspace{-10pt}
\end{figure}

% Daniel: references.

\end{document}